\documentclass{article}  
\usepackage{graphicx}
\usepackage{latexsym}

\usepackage{authblk}

\def\bv{\mathbf{v}}
\def\be{\mathbf{e}}
\def\bu{\mathbf{u}}

\usepackage{tikz}
\usepackage{pgfplots}
\usetikzlibrary{plotmarks,shapes,snakes,pgfplots.dateplot}
\usetikzlibrary{calc,trees,positioning,arrows,chains,shapes.geometric,%
    decorations.pathreplacing,decorations.pathmorphing,shapes,%
    matrix,shapes.symbols,shapes.geometric}
\usetikzlibrary{3d,decorations.text,shapes.arrows,fit,backgrounds,quotes,arrows.meta}

\pgfdeclarelayer{background}
\pgfdeclarelayer{foreground}
\pgfsetlayers{background,main,foreground}

\tikzset{every picture/.style={semithick},every path/.style={thick,rounded corners,->}}

\tikzset{
  ne/.style={
    draw=none, fill=none,
    font=\footnotesize\sffamily, 
    minimum height=0em,
    text centered},
  ncirc/.style={
    circle,
    draw=black, fill=none,
    font=\footnotesize\sffamily, 
    minimum height=2em,
    inner sep=0,
    text centered},
  nd/.style={
    font=\sffamily, 
    text centered},
  nodecomp/.style={
    rectangle, 
    rounded corners, 
    draw=black, thick,
    text width=2em, 
    font=\footnotesize\sffamily,
    minimum height=1.3em, 
    text centered},
  nodevar/.style={
    nodecomp,
    fill=green!10,
  },
  diablo/.style={
    rectangle, 
    rounded corners, 
    draw=black, thick,
    text width=10em, 
    font=\footnotesize\sffamily,
    minimum height=3em, 
    text centered},
  branch/.style ={circle,inner sep=0pt,minimum size=1.5mm,fill=black,draw=black},
  diablo2/.style={
    rectangle, 
    rounded corners, 
    fill=red!10,
    draw=black!80,thick,
    text width=13em, 
    font=\footnotesize\sffamily,
    minimum height=1.5em, 
    text centered},
  diafeature/.style={
    rectangle,
    rounded corners=2pt, 
    fill=green!10,
    draw=black!80,thick,
    text width=1em, 
    font=\footnotesize\sffamily,
    minimum height=6em, 
    text centered},
  diafeatnarr/.style={
    rectangle,
    rounded corners=2pt, 
    fill=green!10,
    draw=black!80,thick,
    text width=.5em, 
    font=\footnotesize\sffamily,
    minimum height=6em, 
    text centered},
  dialoss/.style={
    diablo2,
    fill=green!10,
  },
  rotnode/.style={
    anchor=center, rotate=90, font=\footnotesize\sffamily
  },
  diaext/.style={
    diablo2,
    fill=yellow!40,
  },
  diablo3/.style={
    rectangle, 
    rounded corners, 
    fill=blue!10,
    draw=blue!40,thick,
    text width=3.5em, 
    font=\footnotesize\sffamily\bfseries,
    text=blue,
    minimum height=1.5em,
    text centered},
  line/.style={draw=red,rounded corners,thick, ->, decoration={markings,mark=at position 1 with %
    {\arrow[scale=4,>=stealth]{>}}},postaction={decorate}},
  element/.style={
    tape,
    top color=white,
    bottom color=blue!50!black!60!,
    minimum width=8em,
    draw=blue!40!black!90, very thick,
    text width=10em, 
    minimum height=3.5em, 
    text centered, 
    on chain},
  every join/.style={->,rounded corners,thick,shorten >=1pt},
  decoration={brace},
  lineblue/.style={
  	join,line width=.07cm,->,blue!20
  }
}

\tikzset{pics/fake box/.style args={
#1 with dimensions #2 and #3 and #4}{
code={
  \draw[gray,ultra thin,rounded corners=0pt,fill=#1]  (0,0,0) coordinate(-front-bottom-left) to
  ++ (0,#3,0) coordinate(-front-top-right) --++
  (#2,0,0) coordinate(-front-top-right) --++ (0,-#3,0) 
  coordinate(-front-bottom-right) -- cycle;
  \draw[gray,ultra thin,rounded corners=0pt,fill=#1] (0,#3,0)  --++ 
   (0,0,#4) coordinate(-back-top-left) --++ (#2,0,0) 
   coordinate(-back-top-right) --++ (0,0,-#4)  -- cycle;
  \draw[gray,ultra thin,rounded corners=0pt,fill=#1!80!black] (#2,0,0) --++ (0,0,#4) coordinate(-back-bottom-right)
  --++ (0,#3,0) --++ (0,0,-#4) -- cycle;
}
}}
\tikzset{circle dotted/.style={dash pattern=on .05mm off 2mm, line cap=round}}

\usepackage{paralist}
\usepackage{url}
\usepackage{setspace}
\usepackage{longtable}
\usepackage{caption}
\usepackage{cite}
\usepackage{multirow}
\usepackage{amsmath,amssymb}
\usepackage{booktabs}

\begin{document}

\title{Machine Learning for SAT: \\ Restricted Heuristics and New Graph Representations}

\author[2]{Mikhail Shirokikh}
\author[1]{Ilya Shenbin}
\author[1,2]{\authorcr Anton Alekseev}
\author[1,2]{Sergey Nikolenko}
\affil[1]{Steklov Institute of Mathematics at St. Petersburg, \newline St. Petersburg, Russia}
\affil[2]{St. Petersburg State University, St. Petersburg, Russia}

\maketitle


\begin{abstract}
Boolean satisfiability (SAT) is a fundamental NP-complete problem with many applications, including automated planning and scheduling. To solve large instances, SAT solvers have to rely on heuristics, e.g., choosing a branching variable in DPLL and CDCL solvers. Such heuristics can be improved with machine learning (ML) models; they can reduce the number of steps but usually hinder the running time because useful models are relatively large and slow. We suggest the strategy of making a few initial steps with a trained ML model and then releasing control to classical heuristics; this simplifies cold start for SAT solving and can decrease both the number of steps and overall runtime, but requires a separate decision of when to release control to the solver. Moreover, we introduce a modification of \emph{Graph-Q-SAT} tailored to SAT problems converted from other domains, e.g., open shop scheduling problems. We validate the feasibility of our approach with random and industrial SAT problems. 
\end{abstract}

\section{Introduction}\label{sec:introduction}
Boolean satisfiability (SAT), i.e., deciding whether a Boolean formula in conjunctive normal form (CNF) is satisfiable, is the archetypal NP-complete problem, with numerous applications in computer science. There exist several different classes of approaches to solving SAT: stochastic local search (SLS), Davis--Putman--Logemann--Loveland solvers (DPLL), conflict-driven clause learning (CDCL), ordered binary decision diagrams (OBDD), and others~\cite{10.1561/2200000081}. All known methods are, naturally, exponential in the worst case, and to solve large problems in practice they need to use various heuristics that have a crucial impact on performance: SLS solvers choose a variable to flip, DPLL and CDCL solvers choose branching variables and their assignments, OBDD-based solvers choose clauses for conjunction and variables for projections. In all cases, key decisions have to be made under uncertainty, which opens up possibilities for using machine learning (ML) techniques to improve SAT and SMT solvers.

Various ML-based techniques have arisen to make better heuristic decisions (see ``Related work''). However, they introduce another tradeoff: ML models are usually computationally heavy, and benefits in terms of the number of solver iterations have to be weighted against extra costs the model incurs. Here, we consider \emph{Graph-Q-SAT}~\cite{kurin2020can}, a model that can reduce the number of iterations for a CDCL solver but that is based on a reinforcement learning (RL) agent with a complex graph neural network (GNN) inside.

In this work, we propose a technique for finding this tradeoff: we suggest to make only a few initial steps with a trained RL agent and then release control to classical heuristics. We consider several modifications of this idea: constant number of ``heavy'' steps, a separate action in the RL agent trained to release control, and a separate head in the model to make this decision. We introduce a novel modification with an action pool that uses the (heavy) RL agent once to predict several actions that can be used for several steps. Moreover, we propose a new approach tailored specifically for SAT instances originating from other optimization problems such as open shop scheduling (OSSP). Namely, we construct a GNN whose graph corresponds not to the SAT formula but to the original OSSP instance; this greatly reduces the size of the graph and improves the results in SAT solving. We validate our approaches with an evaluation study on both random and industrial SAT instances.

The paper is organized as follows. Section~\ref{sec:related} surveys related work on machine learning approaches to SAT. Section~\ref{sec:method} introduces \emph{Graph-Q-SAT} (Section~\ref{sec:graphqsat}) and our novel modifications for it (Section~\ref{sec:modifications}). Section~\ref{sec:experiments} reports the experimental setup and datasets and specifies several novel implementation modifications that further improve \emph{Graph-Q-SAT}. Section~\ref{sec:results} presents and discusses the results of our evaluation study, and Section~\ref{sec:conclusion} concludes the paper.

\section{Related work}\label{sec:related}
SAT is a classification problem; end-to-end machine learning (ML) approaches have been proposed by Devlin and O'Sullivan~\cite{devlin2008satisfiability} and with graph neural networks (GNN) by Selsam et al.~\cite{neurosat}, a method later extended to Circuit-SAT by Amizadeh et al.~\cite{amizadeh2018learning} and MAX-SAT by Liu et al.~\cite{DBLP:journals/corr/abs-2111-07568} and improved with recurrent neural networks by Ozolins et al.~\cite{ozolins2021goal}. However, while a satisfying assignment can be verified, end-to-end solutions for an exact combinatorial problem via statistical methods cannot replace a complete solver, i.e., cannot make sure that a formula is unsatisfiable.

Instead of an end-to-end replacement, ML can be used to augment complete SAT solvers by enhancing their \emph{heuristics}. SAT solvers--Davis-Putnam-Logeman-Loveland (DPLL), conflict-driven clause learning (CDCL), and stochastic local search (SLS)---have decisions that have to be made during operation with incomplete information, usually recursively. Such decisions can be improved with ML. Boolean variables initialization has been improved by Wu et al.~\cite{wu2017improving}; in SLS solvers, variable re-initialization is called upon every restart, so Zhang et al.~\cite{zhang2020nlocalsat} train a neural network to improve SLS initial assignments. Nejati et al.~\cite{nejati2017adaptive} learned restart strategies for CDCL solvers while Han et al.~\cite{han2020enhancing} predicted glue variables for the \emph{CaDiCaL} solver previously proposed by Biere et al.~\cite{BiereFazekasFleuryHeisinger-SAT-Competition-2020-solvers}.

The most important direction here is \emph{variable selection} for branching in DPLL and CDCL solvers and flipping in SLS solvers. For DPLL and CDCL, the goal is to improve over the \emph{Variable State Independent Decaying Sum} (VSIDS) heuristic originally introduced in the \emph{Chaff} solver by Moskewicz et al.~\cite{moskewicz2001chaff} that favors recently learned clauses based on specially introduced activity scores; Biere and Fr{\"o}lich~\cite{10.1007/978-3-319-24318-4_29} survey VSIDS variations. Selsam et al.~\cite{selsam2019guiding} learn to predict the unsatisfiable core of a formula, using predictions to re-rank variables for branching; Jaszczur et al.~\cite{jaszczur2020neural} use a GNN in a similar way, while Han~\cite{han2020learning} prioritizes variables that often occur in DRAT proofs, learning to predict this fact.

However, the main downside of ML heuristics for solving SAT is the trade-off between reducing the number of steps and the time required to make predictions with the help of the heavyweight modern machine learning models. Therefore, while most works report significant reductions in the number of iterations from the use of ML-improved heuristics, one should also track the actual running time of the solver with and without these heuristics on equivalent hardware, and most ML methods have limited usability in practice due to computational inefficiency: a heavy neural network has to be used on every branching step.
\emph{NeuroComb} proposed by Wang et al.~\cite{wang2021neurocomb} uses GNNs only for preprocessing, running inference once per input formula (\emph{static} predictions). Many works use reinforcement learning (RL)~\cite{sutton2018reinforcement} to learn heuristics based on the running time or number of iterations as a reward. In particular, Yolcu and P{\'o}czos~\cite{yolcu2019learning} use RL for SLS solvers to choose flip variables, Han~\cite{han2020enhancing} extends \emph{NeuroCore} with RL, Liang et al.~\cite{liang2016exponential,liang2016learning} apply multi-armed bandits as branching heuristics, Lagoudakis and Littman~\cite{lagoudakis2001learning} use TD-learning to select a branching heuristic given the current state, while Nejati et al.~\cite{nejati2017adaptive} apply multi-armed bandits to choose a restart strategy.

\emph{Graph-Q-SAT} by Kurin et al.~\cite{kurin2020can} utilizes Q-learning for branching heuristics in CDCL solvers via a Markov decision process (MDP) whose states are defined by unassigned variables and unsatisfied clauses with these variables; we will describe this approach in detail below. We also note here that Wang et al.~\cite{wang2018gameplay} already introduced a similar MDP formulation, which did not use GNNs and instead was based on \emph{AlphaGo Zero}. Interestingly, games such as Go and chess indeed give rise to computationally intractable problems, e.g., the game of Go on an $n\times n$ board has been shown to be EXPTIME-complete~\cite{DBLP:conf/ifip/Robson83} and chess to be PSPACE-complete~\cite{STORER198377}, so these two directions of research are not as different as it might seem.

\section{Proposed method}\label{sec:method}
\subsection{Graph-Q-SAT}\label{sec:graphqsat}

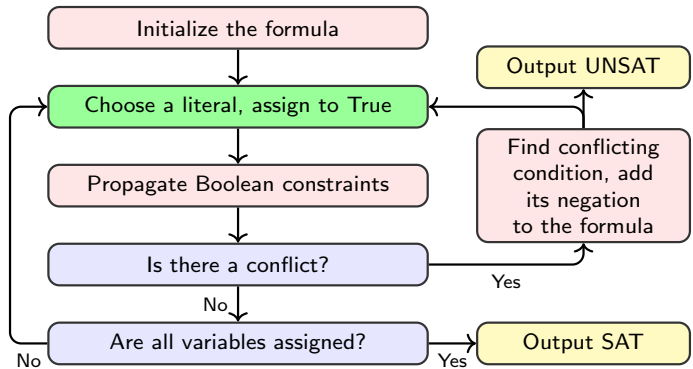
\begin{figure}[!t]\centering
\resizebox{.75\linewidth}{!}{\begin{tikzpicture}[node distance=.3cm]
\def\xx{1.75}
\def\yy{1}
    \node[diablo2] (a1) at (0*\xx,5*\yy) {Initialize the formula};
    \node[diablo2, fill=green!40] (a2) at (0*\xx,4*\yy) {Choose a literal, assign to True};
    \node[diablo2] (a3) at (0*\xx,3*\yy) {Propagate Boolean constraints};
    \node[diablo2, fill=blue!10] (a4) at (0*\xx,2*\yy) {Is there a conflict?};
    \node[diablo2, fill=blue!10] (a5) at (0*\xx,1*\yy) {Are all variables assigned?};
    \node[diablo2, text width=7em] (b1) at (2.5*\xx,3*\yy) {Find conflicting condition, add its negation to the formula};
    \node[diablo2, text width=7em, fill=yellow!30] (b2) at (2.5*\xx,1*\yy) {Output SAT};
    \node[diablo2, text width=7em, fill=yellow!30] (b3) at (2.5*\xx,4.5*\yy) {Output UNSAT};
    \draw (a1) -- (a2);
    \draw (a2) -- (a3);
    \draw (a3) -- (a4);
    \draw (a4) -- node[midway,left] {\sffamily\scriptsize No} (a5);
    \draw (a4) -| node[near start,below] {\sffamily\scriptsize Yes} (b1);
    \draw (a5) -- node[midway,below] {\sffamily\scriptsize No} ++(-1.65*\xx, 0) |- (a2);
    \draw (a5) -- node[midway,below] {\sffamily\scriptsize Yes} (b2);
    \draw (b1) -- (b3);
    \draw (b1) |- (a2);
\end{tikzpicture}}

\caption{CDCL solver operation: choosing an assignment is a heuristic step that can be improved by machine learning.}\label{fig:cdcl}
\end{figure}

In this work, we build upon the \emph{Graph-Q-SAT} approach by Kurin et al.~\cite{kurin2020can} that uses Q-learning for branching heuristics in CDCL solvers~\cite{series/faia/SilvaLM09}. A CDCL solver operates as shown in Fig.~\ref{fig:cdcl}: on every step, it
\begin{itemize}
    \item chooses a literal (variable and its value) to assign, 
    \item then propagates Boolean constraints such as, e.g., unit clauses (clauses with one literal) that lead to more necessary assignments.
\end{itemize}

If the solver arrives at a conflict (empty clause), it finds the conflicting condition (a cut in the implication graph), backtracks to the assignment where it occurred, and adds the clause representing its negation to the formula. If the algorithm has backtracked from both possible assignments of a variable, the formula is unsatisfiable. If all variables have been assigned with no conflict, the solver has found a satisfying assignment.

On every iteration, a CDCL solver has to choose a literal. This part is a heuristic, and choosing correct variables can lead to exponential speedups in the running time and number of iterations needed for both SAT and UNSAT instances, so this is the place where ML-based heuristics may help. 

\emph{Graph-Q-SAT} is a reinforcement learning method that learns to choose literals by treating the problem as a Markov decision process (MDP):
\begin{itemize}
    \item 
    the task 
    $\tau \sim D(\phi, (\mathrm{un})\mathrm{SAT}, n_{\mathrm{vars}}, n_{\mathrm{clauses}})$ 
    is a SAT problem,
    sampled from a task family $\phi$ with known satisfiability status and given number of variables and clauses;
    \item 
    each state of the MDP is defined by unassigned variables and unsatisfied clauses with these variables;
    \item 
    actions available to the agent are the choice of a variable and its polarity;
    \item
    a terminal state is reached when either a satisfying assignment or an unsat core is found, and the final reward is $1$ with a discount factor $\gamma < 1$ to reward solving SAT faster.
\end{itemize}

\begin{figure}[!t]\centering\setlength{\tabcolsep}{15pt}
\begin{tabular}{cc}
\begin{tikzpicture}[node distance=.3cm]
\def\xx{2}
    \node[ncirc, fill=blue!10] (x1) at (0*\xx,2) {$x_1$};
    \node[ncirc, fill=blue!10] (x2) at (0*\xx,1) {$x_2$};
    \node[ncirc, fill=blue!10] (x3) at (0*\xx,0) {$x_3$};
    \node[ncirc, fill=blue!10] (x4) at (0*\xx,-1) {$x_4$};

    \node[ncirc, fill=red!10] (c1) at (1*\xx,1.5) {$c_1$};
    \node[ncirc, fill=red!10] (c2) at (1*\xx,-0.5) {$c_2$};

    \draw[-] (x1) -- node[midway,above] {\scriptsize $[1,0]$} (c1);
    \draw[-] (x2) -- node[midway,above] {\scriptsize $[1,0]$} (c1);
    \draw[-] (x3) -- node[midway,right] {\scriptsize $[0,1]$} (c1);
    \draw[-] (x2) -- node[midway,right] {\scriptsize $[0,1]$} (c2);
    \draw[-] (x3) -- node[near start,below] {\scriptsize $[1,0]$} (c2);
    \draw[-] (x4) -- node[midway,below] {\scriptsize $[1,0]$} (c2);
\end{tikzpicture} 
& 
\begin{tikzpicture}[node distance=.3cm]
\def\xx{2}
    \node[ncirc, fill=blue!10] (x1) at (0*\xx,2) {$x_1$};
    \node[ncirc, fill=blue!10] (x2) at (0*\xx,1) {$x_2$};
    \node[ncirc, fill=blue!10] (x3) at (0*\xx,0) {$x_3$};
    \node[ncirc, fill=blue!10] (x4) at (0*\xx,-1) {$x_4$};
    \node[ncirc, fill=red!10] (cc1) at (1*\xx,2) {$c_1$};
    \node[ncirc, fill=green!10] (c1) at (1*\xx,1) {$\rho_{e\to v}$};
    \node[ncirc, fill=green!10] (v) at (1.75*\xx,1) {$\phi_v$};
    \node[ncirc, fill=red!10] (cc2) at (1*\xx,-1) {$c_2$};
    \node[ncirc, fill=green!10] (e) at (1.25*\xx,0) {$\phi_e$};
    \node[ncirc, fill=yellow!30] (u) at (2*\xx,-1) {$\bu$};

    \draw (x1) -- node[near end,above] {\scriptsize $\be_{11}$} (c1);
    \draw (x2) -- node[midway,above] {\scriptsize $\be_{21}$} (c1);
    \draw (x3) -- node[midway,right] {\scriptsize $\be_{31}$} (c1);
    \draw (c1) -- (v);
    \draw (u) -- (v);
    \draw (u) -- (e);
    \draw (cc2) -- node[midway,right] {\scriptsize $\bv_{c2}$} (e);
    \draw (x4) -- node[midway,above] {\scriptsize $\bv_{x4}$} (e);
    \draw (cc1) -- node[midway,above] {\scriptsize $\bv_{c1}$} (v);
\end{tikzpicture}
\\ (a) & (b) \end{tabular}

\caption{\emph{Graph-Q-SAT}: (a) the variable-clause graph for the formula $(x_1\lor x_2\lor\lnot x_3)\land (\lnot x_2 \lor x_3\lor x_4)$; (b) computation of sample GNN updates for $\bv_{c1}$ (top) and $\be_{x4,c2}$ (bottom).}\label{fig:gnn}
\end{figure}
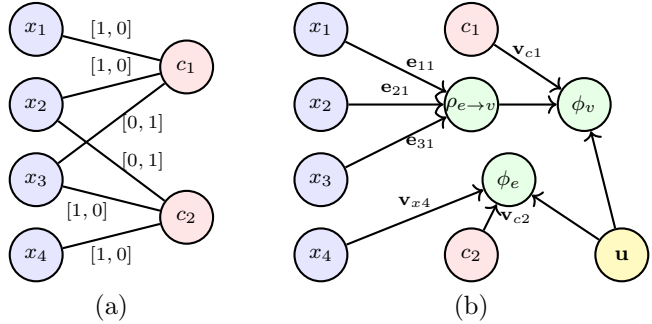

This is a sufficiently general formulation that has been only slightly modified in other works that apply RL to SAT. \emph{Graph-Q-SAT} uses \emph{Q-learning} that learns to predict expected utilities for each action; for SAT, it means that for a given formula, \emph{Graph-Q-SAT} learns to predict expected utilities $Q(l)$ (running time rewards) based on the formula structure for each literal $l$; we can choose $\arg\max_lQ(l)$ as the next literal to assign during inference. See, e.g., Mnih et al.~\cite{DBLP:journals/corr/MnihKSGAWR13} for a detailed discussion of (deep) Q-learning.

To estimate $Q(l)$, \emph{Graph-Q-SAT} uses a \emph{graph neural network} (GNN)~\cite{DBLP:journals/corr/abs-1806-01261}, which is a common theme in ML for SAT~\cite{neurosat,yolcu2019learning}. It uses the variable-clause graph to represent formulas, i.e., a formula with $n$ variables and $m$ clauses corresponds to a bipartite graph $G=(V, E)$ with $n$ and $m$ variables in each part (see Fig.~\ref{fig:gnn}a). An edge $(x_i, c_j)\in E$ means that $x_i$ occurs in the clause $c_j$; an edge is labeled $[0, 1]$ if $c_j$ contains the positive literal $x_i$ and $[1, 0]$ if it contains $\lnot x_i$.

On each step, GNN propagates information along the edges of $G$, updating the embeddings of edges $\be_{ij}$, embeddings of vertices $\bv_i$, and a global context vector $\bu$. The network is defined by update functions $\phi_e$, $\phi_v$, $\phi_u$ that change the embeddings and aggregation functions $\rho_{e\to v}$, $\rho_{e\to u}$, $\rho_{v\to u}$ that combine the embeddings of adjacent entities. 

Formally, on every iteration the GNN updates (see Fig.~\ref{fig:gnn}b)
\begin{align*}
\be_{ij} :=& \phi_e\left(\bu, \be_{ij}, \bv_i, \bv_j\right)\quad\forall\ e_{ij}\in E,\\
\bv_{i} :=& \phi_v\left(\bu, \bv_i, \rho_{e\to v}\left(\left\{\be_{ki}\mid e_{ki}\in E\right\}\right)\right)\quad\forall\ v_{i}\in V,\\
\bu :=& \phi_u\left(\bu, \rho_{e\to u}\left(\left\{\be\mid e\in E\right\}\right), \rho_{v\to u}\left(\left\{\bv\mid v\in V\right\}\right)\right),
\end{align*}
repeating until convergence or until it is stopped.
Sample computations are illustrated in the graph shown in Fig.~\ref{fig:gnn}b.

The functions are parameterized with neural networks, trained with backpropagation on the learning phase. The original \emph{Graph-Q-SAT} uses the \emph{Encoder}--\emph{Core}--\emph{Decoder} architecture that consists of three significantly different parts. The encoder and decoder do not aggregate along the edges; they apply linear layers with \emph{ReLU} activation functions to the feature vectors $\bv$, $\be$, and $\bu$. The core is located between them and consists of 4 GNN layers as described above. Interestingly, while the original \emph{Graph-Q-SAT} code does have generic encoder and decoder parts implemented, the version introduced by Kurin et al.~\cite{kurin2020can} has no hidden layers in the encoder and decoder.

As the underlying solver \emph{Graph-Q-SAT} uses \emph{MiniSAT}~\cite{minisat}, an award-winning CDCL solver that chooses the next branching literal based on their \emph{activity scores}, with a variation of the \emph{Variable State Independent Decaying Sum} (VSIDS) heuristic; it favors recently learned clauses, increasing activity scores for variables in recent clauses and decaying them with time. \emph{Graph-Q-SAT} replaces VSIDS scores with predicted values of $Q(l)$.

\subsection{Our modifications}\label{sec:modifications}

We have made several modifications and improvements for the \emph{Graph-Q-SAT} framework. First and most importantly, we note that running a graph neural network on every step of SAT solving is a very resource-intensive task; even if it helps further reduce the number of steps a little, it certainly hurts the running time, usually drastically. The most influential decisions come at the beginning, and this is exactly when classical SAT solving heuristics such as MiniSAT activity scores are not yet meaningful (they don't yet have enough statistics). Therefore, we consider two ways of limiting the time:
\begin{enumerate}[(1)]
    \item running \emph{Graph-Q-SAT} for a fixed number of steps and then releasing control to MiniSAT;
    \item learning when to release control to MiniSAT, i.e., training a separate head that predicts this.
\end{enumerate}
We also experimented with running the model once per several steps, but these experiments did not bring improvements and are not reported below.

\begin{figure}[!t]\centering
\begin{tabular}{c}
\resizebox{.8\linewidth}{!}{\begin{tikzpicture}[node distance=.3cm]
\def\xx{4}
\def\yy{1}
    \node[diablo2, text width=4em, fill=blue!10] (a1) at (0*\xx,5*\yy) {Solver state};
    \node[diablo2, text width=4em, fill=blue!10] (a2) at (1*\xx,5*\yy) {Clause-variable graph};
    \node[diablo2, text width=4em, fill=blue!10] (a3) at (2*\xx,5*\yy) {Q-values};
    \draw[red!80!white] (a1) -- node[midway,above] {\sffamily\footnotesize Gym env.} (a2);
    \draw[red!80!white] (a2) -- node[midway,above] {\sffamily\footnotesize GNN} (a3);

    \node[diablo2, text width=4em, fill=blue!10] (a4) at (1.5*\xx,3*\yy) {Action};
    \node[diablo2, text width=5em, fill=blue!10] (a5) at (0.5*\xx,3*\yy) {Variable assignment};

    \draw[green!50!black] (a3) -- node[midway,right] {\sffamily\footnotesize Agent} (a4);
    \draw[green!50!black] (a4) -- node[midway,above] {\sffamily\footnotesize Gym env.} (a5);
    \draw[red!80!white] (a5) -- node[midway,left] {\sffamily\footnotesize Minisat} (a1);
\end{tikzpicture}}
\\ (a) \\[5pt]
\resizebox{.8\linewidth}{!}{\begin{tikzpicture}[node distance=.3cm]
\def\xx{4}
\def\yy{1}
    \node[diablo2, text width=4em, fill=blue!10] (a1) at (0*\xx,5*\yy) {Solver state};
    \node[diablo2, text width=3em, fill=green!40] (b1) at (0.5*\xx,5*\yy) {Action pool empty?};
    \node[diablo2, text width=4em, fill=blue!10] (a2) at (1.25*\xx,5*\yy) {Clause-variable graph};
    \node[diablo2, text width=4em, fill=blue!10] (a3) at (2*\xx,5*\yy) {Q-values};
    \draw[green!50!black] (a1) -- (b1);
    \draw[red!80!white] (b1) -- node[midway,above] {\sffamily\footnotesize Yes} (a2);
    \draw[red!80!white] (a2) -- node[midway,above] {\sffamily\footnotesize GNN} (a3);
    
    \node[diablo2, text width=5em, fill=blue!10] (b2) at (1.5*\xx,3*\yy) {Action pool:\\ Action 1 \\$\ldots$ \\ Action $k$};

    \node[diablo2, text width=5em, fill=blue!10] (a5) at (0.5*\xx,3*\yy) {Variable assignment};

    \draw[green!50!black] (a3) -- node[midway,right] {\sffamily\footnotesize Agent} (b2);
    \draw[green!50!black] (b2) -- node[midway,below] {\sffamily\footnotesize Gym env.} (a5);
    \draw[red!80!white] (a5) -- node[midway,left] {\sffamily\footnotesize Minisat} (a1);
    \draw[green!50!black] (b1) -- node[midway,above] {\sffamily\footnotesize No} (b2);
\end{tikzpicture}}
\\ (b)
\end{tabular}

\caption{Using the action pool: (a) regular \emph{Graph-Q-SAT} operation; (b) our modification with the action pool. Green arrows show ``fast'' operations; red arrows, ``slow'' operations that incur significant computational costs.}\label{fig:pool}
\end{figure}
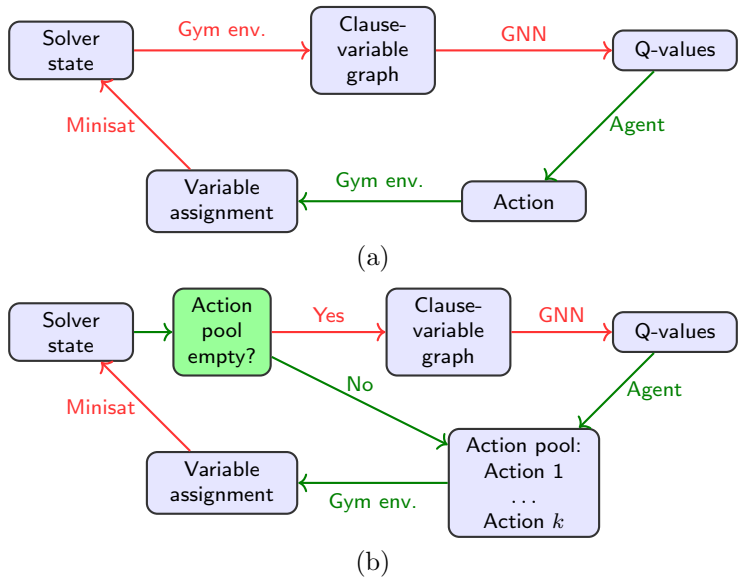

Second, we propose one more way to reduce the number of calls to the agent, namely to utilize every run of the agent network in a more efficient way. The graph that represents the current formula changes only slightly after each step of the SAT solving agent. Therefore, we can assume that the list of actions with the highest Q-values in most cases will also change only slightly, which is confirmed by experiments. Following this motivation, we propose to keep top $k$ actions that correspond to the highest Q-values and perform these actions sequentially (ignoring actions that have become invalid by the time of their execution). Such a policy is used at test time only, not at train time. The number $k$, i.e., the number of actions per agent run, is a constant hyperparameter, although we note as a possible extension for future work that the number $k$ could also be adaptive. The value of $k$ should be large enough to get sufficient runtime acceleration in SAT solving while remaining small enough so that the environment does not change too much and the values of the $Q$ function remain relevant. We call this set of precomputed actions the \emph{action pool}; the resulting workflow is shown in Fig.~\ref{fig:pool}b in comparison with the regular \emph{Graph-Q-SAT} workflow (Fig.~\ref{fig:pool}a).

Third, we have experimented with \emph{Graph-Q-SAT} architectures. In the final version, we:
\begin{itemize}
    \item turned on the encoder and decoder parts;
    \item increased the number of core layers from 4 to 13;
    \item increased the depth of core layers from 1 to 2;
    \item added a separate ``release control'' action predicted with two linear layers over the penultimate layer of the network.
\end{itemize}

Third, Kurin et al.~\cite{kurin2020can} use a MiniSAT implementation that serves as the RL environment for \emph{Graph-Q-SAT}, and this significantly increases the runtime. Since we release control to MiniSAT and do not return to RL, we are able to use the original MiniSAT after release, which has sped up SAT solving by a factor of about 10x; in all experiments, we report the runtimes of our version.

\section{Experimental setup}\label{sec:experiments}
\subsection{Model setup and training}

Similar to Kurin et al.~\cite{kurin2020can}, we used Random 3-SAT instances during training, with $50$ variables and $218$ clauses each (close to the phase transition threshold), and coloring problems for random graphs with $75$ vertices and $180$ edges; see also the SATLIB benchmark~\cite{satlib}.

Since we need to learn to switch between RL and VSIDS heuristics, we have changed the rewards. With constant $-p$ reward for every step, as in the original Graph-Q-SAT~\cite{kurin2020can}, the agent would have no motivation to flip the switch as long as his variable assignments are better than random. Therefore, instead of a constant penalty $-p$ for every iteration we used $-t$, where $t$ is the CPU time in milliseconds for this step (higher for RL and lower for VSIDS).

We also turned off discounting for $Q$ values (set $\gamma=1$) to motivate the agent to look ahead to long-term costs and switched to Double DQN instead of regular DQN; both of these changes significantly improved the agent's behaviour in our experiments.

We have also expanded the action space with a new one that releases control to VSIDS. During training, we force the agent to make a small constant number of steps $n$ since otherwise for an untrained agent this action would always be preferable; the value of $n$ did not have a significant effect on the results. We also replaced DQN learning with Double DQN~\cite{DBLP:journals/corr/HasseltGS15}, which stabilized the training process.

Note that the reward received after the switch varies over time, which does not match the conventional definition of MDPs. However, we can treat this model as an MDP as follows: suppose that at time $t$ we have a valid action with reward $r_t$ that switches control to VSIDS, and at time $t+1$ we have a different action that does the same but obtains reward $r_{t+1}$. In the implementation, it turns out that parameters of the action with reward $r_{t+1}$ are initialized by parameters of the action with reward $r_t$.

Each model is run three times on every task, and the tables show averaged results.

\subsection{Evaluation datasets}

For evaluation, we have chosen several traditional SAT datasets and generated open shop scheduling problems (OSSP).

\paragraph*{$\mathbf{SR}(n)$.} For random generation, we use the $\mathbf{SR}(n)$ distribution introduced by Selsam et al.~\cite{neurosat}. To generate a random clause on $n$ variables, $\mathbf{SR}(n)$ samples a small integer $k$ with mean a little over $4$, namely as a sum of two random variables 
$$1 + \mathbf{Bernoulli}(0.7) + \mathbf{Geo}(0.4)$$ 
(since there should not be many $2$-literal clauses), then samples $k$ variables uniformly at random without replacement, and negates each one with probability $0.5$. 

The clauses $c_i$ are consecutively generated and added to the SAT instance until the problem becomes unsatisfiable at step $m$. Since we know that it is only $c_m$ that ``breaks'' satisfiability, negating any literal in $c_m$ leads to a satisfiable formula with clauses $\{c_1, ..., c_{m-1}, c'_m\}$; both $c_1 \land ... \land c_{m-1}\land c'_m$ and $c_1 \land ... \land c_m$ are added to the set. This is a standard distribution for SAT evaluation~\cite{neurosat,yolcu2019learning}.

\paragraph*{Industrial.} This is the \emph{maris05} subset of industrial problems from the SAT 2005 Competition benchmark\footnote{Available at \url{https://zenodo.org/record/6528885}}.

\paragraph*{OSSP datasets.}
 Open shop scheduling is a standard scheduling problem; the task is to distribute the workload across a finite number of resources (machines). The workload is represented as a list of jobs; a job is essentially a set of order-independent operations that each must be processed on a certain machine, i.e., we need to schedule $n$ jobs on $m$ machines, where each of the $n$ jobs has its own sequence of operations that have to be done within $m$ machines, with the goal to optimize an objective function under constraints that can define termination times, delay times, or total flow times, among others. For each job, only one operation can be run at a given moment of time (i.e., jobs cannot be run ``in parallel''), and each machine can be occupied with only one operation at a time. The timespans $p_*$ required for each operation to be completed are given. The open shop scheduling problem (OSSP) is to find the schedule with the shortest \textit{makespan}, i.e., the time interval between the start of work and the end of the last completed job. For $3$ or more machines and $3$ or more jobs, OSSP is known to be NP-hard~\cite{williamson1997short}.

We have generated a problem set proposed by Taillard~\cite{taillard1993benchmarks}\footnote{The generator (Pascal code) is available, e.g., at \url{http://people.brunel.ac.uk/~mastjjb/jeb/orlib/files/openshop.txt}}; upper and lower bounds for the makespans are provided. For default conversion to CNF, we have chosen the Crawford-Baker encoding~\cite{crawford1994experimental}.

Specifically, the Crawford-Baker encoding enumerates all operations (a batch of work for a single job that has to be processed on a certain machine) and introduces the following Boolean \emph{variables}:
\begin{itemize}
    \item $sa_{i,t}$: ``op $i$ starts at time $t$ or later'',
    \item $eb_{i,t}$: ``op $i$ ends by time $t$'',
    \item $pr_{i,j}$: ``op $i$ precedes $j$''.
\end{itemize}
The \emph{constraints} are formulated as follows:
\begin{itemize}
    \item $\lor_t sa_{i,t}$ for all operations $i$ (each operation begins);
    \item $\lor_t eb_{i,t},~t>0$ for all operations $i$ (each operation ends);
    \item $pr_{i,j} \lor pr_{j,i}$ for all operations $i$, $j$ from the same job that are run on the same machine (they cannot run at the same time);
    \item $sa_{i,t} \to sa_{i,t-1}$ (an implication $a \to b$ is the disjunct $\lnot a \lor b$) for all operations and all times $t$: if $i$ starts at or after $t$, it starts at or after $t-1$ ($sa$ coherence);
    \item $eb_{i,t} \to eb_{i,t+1}$ for all possible $i$ and $t$ ($eb$ coherence);
    \item $sa_{i,t} \to \lnot eb_{i,t+p_i-1}$ for all $i$ and $t$, where $p_i$ is the processing time (operation $i$ that started at time $t$ or later cannot end before $t + p_i$);
    \item $sa_{i,t} \land pr_{i,j} \to sa_{j,t+p_i}$ for all $i$, $j$, and times $t$ (operation $j$ cannot start until operation $i$ ends if $i$ precedes $j$).
\end{itemize}

\subsection{Graph-Q-SAT implementation modifications}\label{sec:implementation}

We have made several modifications and improvements to the original \emph{Graph-Q-SAT} implementation. First, the Graph-Q-SAT paper~\cite{kurin2020can} makes comparisons against not the original \emph{Minisat} solver but against {\tt MinisatAgent}, an RL environment agent acting as \emph{Minisat}. The number of steps remains the same, but the running time required for obtaining the results via {\tt MinisatAgent} is significantly increased compared to the original \emph{Minisat} solver.

To make our approach more feasible for real-life scenarios, we have added a new action to the environment (trainable in the \emph{CoPilot} setting); when this action is triggered, the environment runs the original \emph{Minisat} without rebuilding the graph or any calls to \emph{Python} code whatsoever, which reduces runtimes very significantly.

In the case of choosing several sequential actions at once, the original \emph{Graph-Q-SAT} implementation of both agent and environment resulted in the most time-consuming stage of each of the agent-environment interaction to become graph construction. But when we apply several predefined actions one by one, graph construction for every action is no longer necessary, and actions can be simply pulled from the action pool; thus, we made the action pool a part of the environment to avoid unnecessary computations.

\section{Evaluation results}\label{sec:results}
\def\gqorig{GQSAT}
\def\gq{GQSAT-2}
\def\ms{MiniSAT}
\def\msnr{MiniSAT without restarts}
\def\cop{CoPilot}
\def\gm{GQSAT-multistep}

\begin{table*}[!t]\centering
\setlength{\tabcolsep}{4pt}
\caption{Evaluation on standard SAT benchmarks.}\label{tbl:eval}
\begin{tabular}{lcc|rr|rr|rr}\toprule
 & {\footnotesize\# of} & {\footnotesize Q-Activity} & \multicolumn{2}{c|}{$\mathbf{SR}(300)$} & \multicolumn{2}{c|}{$\mathbf{SR}(500)$} & \multicolumn{2}{c}{\textbf{Industrial}} \\
& {\footnotesize model steps} & & Time & Steps & Time & Steps & Time & Steps \\
\midrule
\ms & & & 0.069 & 655.1 & 1.292 & 14317.4 & 0.386 & 1033.9 \\
\midrule
\gqorig & 1 & & 0.096 & 656.9 & 1.270 & 13770.4 & 0.550 & 1220.2 \\
 & 2 & & 0.121 & 671.0 & 1.377 & 14294.4 & 0.516 & 991.0 \\
 & 3 & & 0.145 & 680.4 & 1.371 & 13918.8 & 0.597 & 1030.8 \\
 & 5 & & 0.180 & 640.7 & 1.429 & 13856.2 & 0.745 & 1070.8 \\
 \midrule
\gq & 1 & & 0.119 & 644.6 & 1.425 & 14804.4 & 0.512 & 981.4 \\
 & 2 & & 0.160 & 645.6 & 1.486 & 14642.7 & 0.628 & 882.9 \\
 & 3 & & 0.203 & 654.7 & 1.693 & 15791.8 & 0.798 & 1074.3 \\
 & 5 & & 0.278 & 661.3 & 1.511 & 13099.7 & 1.097 & 1318.8 \\
 \midrule
\cop  & 1 & & 0.121 & 644.6 & 1.422 & 14804.4 & 0.558 & 1177.3 \\
 & 2 & & 0.160 & 645.6 & 1.487 & 14642.7 & 0.639 & 986.9 \\
 & 3 & & 0.201 & 654.7 & 1.708 & 15791.8 & 0.794 & 1026.7 \\
 & 5 & & 0.235 & 654.7 & 1.750 & 15791.8 & 0.904 & 1210.0 \\
 \midrule
\cop & 1 & $\checkmark$ & 0.123 & 644.6 & 1.419 & 14804.4 & 0.521 & 1019.3 \\
 & 2 & $\checkmark$ & 0.160 & 645.6 & 1.477 & 14642.7 & 0.630 & 926.9 \\
 & 3 & $\checkmark$ & 0.204 & 654.7 & 1.704 & 15791.8 & 0.797 & 1102.3 \\
 & 5 & $\checkmark$ & 0.231 & 638.8 & 1.516 & 13759.8 & 0.973 & 1217.2 \\
\bottomrule
\end{tabular}\vspace{.4cm}
\end{table*}

\begin{table*}[!t]\centering
\setlength{\tabcolsep}{4pt}
\caption{Evaluation results for the action pool modification.}\label{tbl:eval2}
\begin{tabular}{lcc|rr|rr|rr}\toprule
 & {\footnotesize \# of} & {\footnotesize Action} & \multicolumn{2}{c|}{$\mathbf{SR}(300)$} & \multicolumn{2}{c|}{$\mathbf{SR}(500)$} & \multicolumn{2}{c}{\textbf{Industrial}} \\
 & {\footnotesize model} & {\footnotesize pool} &  &  &  &  &  &  \\
& {\footnotesize runs} & {\footnotesize size} & Time & Steps & Time & Steps & Time & Steps \\
\midrule
\ms & & & 0.071 & 759.7 & 0.941 & 11292.7 & 1.881 & 4825.8 \\
\midrule
\gqorig & 1 & & 0.088 & 694.1 & 0.944 & 10443.8 & 1.867 & 4828.0 \\
 & 2 & & 0.124 & 611.6 & 0.848 & 9719.1 & 1.931 & 4805.6 \\
 & 3 & & 0.149 & 588.5 & 0.853 & 9507.9 & 2.166 & 4810.9 \\
 \midrule
\gm  & 1 & 20 & 0.091 & 611.8 & 0.724 & 8718.5 & 1.842 & 4813.9 \\
 & 2 & 20 & 0.107 & 571.4 & 0.721 & 8398.3 & 1.898 & 4532.0 \\
 & 3 & 20 & 0.131 & 601.7 & 0.735 & 8409.9 & 1.983 & 4496.9 \\
 \midrule
\gm  & 1 & 30 & 0.089 & 587.4 & 0.715 & 8508.3 & 1.771 & 4668.8 \\
 & 2 & 30 & 0.110 & 600.6 &  0.732 & 8577.3 & 1.976 & 4734.7 \\
 & 3 & 30 & 0.130 & 604.8 & 0.763 & 8681.3 & 3.332 & 7177.7 \\
\bottomrule
\end{tabular}\vspace{.4cm}
\end{table*}

\subsection{Results on standard SAT benchmarks} 

Table~\ref{tbl:eval} shows the main results of our experimental evaluation. In Table~\ref{tbl:eval}:
\begin{itemize}
    \item \gqorig{} is the original \emph{Graph-Q-SAT} that uses the RL agent for $1$, $2$, $3$, or $5$ steps;
    \item \gq{} denotes our extended \emph{Graph-Q-SAT} architecture; both versions use our reimplementation of \emph{Graph-Q-SAT} as described in Section~\ref{sec:implementation};
    \item MiniSAT denotes the \gqorig{} implementation where the RL model is not used;
    \item \cop{} is the \gq{} model with an additional action for releasing control to MiniSAT, still with a bounded number of steps.
\end{itemize}

In all cases, MiniSAT is run with restarts; we report average running times and numbers of iterations. The table shows that restricting the number of RL-based steps can improve the number of steps while saving time compared to running it for a longer time; \cop{} can improve the results further.

Another important idea here is that even if we release control to MiniSAT, we have already spent time on constructing the model and getting its predictions on first steps. We can reuse the estimated $Q$ functions in MiniSAT's activity scores, thus helping it with cold start. For each variable $x_i$, we use $-\frac{1}{\max\left(Q(x_i), Q(\lnot x_i)\right)}$ as the initial value of its activity score, and then they are modified according to usual MiniSAT rules. The column ``Q-activity'' shows this variation of the model,
yielding significant improvements.

\begin{table*}[!t]\centering
\setlength{\tabcolsep}{4pt}
\caption{Evaluation results for the new OSSP representation.}\label{tbl:newgraph}
\begin{tabular}{lc|rr|rr|rr}\toprule
 & {\footnotesize \# steps} & \multicolumn{2}{c|}{\textbf{$5\times 5$ OSSP}} & \multicolumn{2}{c|}{\textbf{$6\times 6$ OSSP}} & \multicolumn{2}{c}{\textbf{$7\times 7$ OSSP}} \\\midrule
\ms  & & 0.886 & 167.4 & 4.483 & 402.7 & 14.444 & 1153.6 \\
\midrule
\gqorig  & 1 & 1.557 & 165.8 & 5.643 & 406.5 & 16.978 & 1152.1 \\
 & 2 & 2.174 & 169.7 & 7.201 & 420.5 & 19.028 & 1149.0 \\
 & 3 & 2.787 & 164.6 & 8.926 & 431.1 & 22.281 & 1144.2 \\
 \midrule
\cop & 1 & 0.890 & 154.7 & 4.423 & 406.8 & 14.423 & 1151.5 \\
with the new & 2 & 1.080 & 170.2 & 4.677 & 403.4 & 14.869 & 1157.0 \\
graph representation & 3 & 1.254 & 171.0 & 5.303 & 387.3 & 16.350 & 1138.9 \\
 \bottomrule
\end{tabular}
\end{table*}

\subsection{ML for SAT with an action pool}

\begin{figure}[!t]\centering\setlength{\tabcolsep}{7pt}
\begin{tabular}{cc}
\includegraphics[width=.45\linewidth]{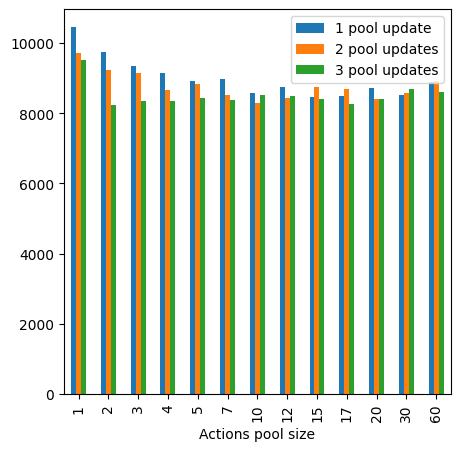} &
\includegraphics[width=.45\linewidth]{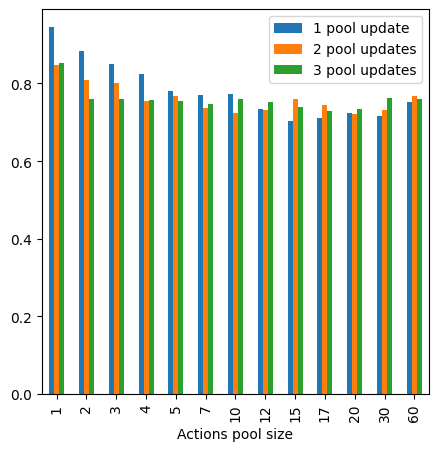} \\
(a) Number of steps &
(b) Running time, s
\end{tabular}

\caption{Experimental results for the SR(500) benchmark.}\label{fig:sr500}
\end{figure}

Table~\ref{tbl:eval2} presents the results for our \emph{Graph-Q-SAT} modification proposed in Section~\ref{sec:modifications}, where the agent chooses several actions at once (Fig.~\ref{fig:pool}b). We show several values of the action pool size (number of ``precomputed'' actions) and numbers of model runs (phrased in this way because now a model run corresponds to several steps); MiniSAT is run without restarts in this table. Figure~\ref{fig:sr500} shows sample more detailed results for the SR(500) benchmark family in terms of both number of model runs (Fig.~\ref{fig:sr500}a) and running time (Fig.~\ref{fig:sr500}b) shown as a function of the action pool size.

Table~\ref{tbl:eval2} shows that the proposed approach with the action pool combines a significant reduction in the number of steps (much more so than in Table~\ref{tbl:eval}) with virtually no loss in evaluation runtime; we note here that the neural networks involved in our experiments were run in their default \emph{PyTorch} implementations without any optimization tricks, and in an industrial environment ML-assisted runtimes might become significantly faster.

\subsection{OSSP modification} 
The OSSP dataset highlights the main weakness of \emph{Graph-Q-SAT}: even the smallest nontrivial (i.e., non-polynomial in the worst case) set, \emph{ossp3-3}, has thousands of variables and clauses, which is not a problem for classical SAT solvers but bloats the \emph{Graph-Q-SAT} model size and makes each RL agent step extremely expensive even during inference, let alone training the agent.

Hence, we propose a novel approach to solving OSSP via SAT with GNN-based heuristics. We still encode OSSP as SAT instances and use MiniSAT with neural heuristics, but we introduce a more compact graph representation for OSSP. 

In our representation, the vertices correspond to operations, and every vertex is labeled with operation time, first possible timestamp when it can begin, and last timestamp when it has to finish (deadline). The first number is constant while the second and third depend on the current solver state; the labels are initialized with $(p_i, 1, T)$ and then are computed from the solver state on every step. We add an edge from vertex $i$ to $j$ if operation $i$ can be done before $j$ and $i$ and $j$ must be done on the same machine or belong to the same job (add edges between vertices whose operations cannot be done at the same time). A variable assignment now corresponds to choosing an edge: an edge from $i$ to $j$ means that operation $i$ must be performed before op $j$, i.e., we set $pr_{ij}=\mathrm{True}$. 

The resulting graph is much smaller than the SAT graph for the Crawford-Baker encoding: for $j$ jobs and $m$ machines it has $jm$ vertices and $O( jm(j+m) )$ edges (note that this size does not depend on $T$) while the SAT formula has $O(j^2m^2 + jmT)$ variables and $O(j^2m^2(j+m) + T(jm)^2 )$ clauses. For instance, a sample $7\times 7$ OSSP task is encoded here with $49$ vertices and $84$ edges, while the SAT formula graph contains over $54$K variables and $1.1$M clauses. In fact, \emph{Graph-Q-SAT} graphs for realistic sized problems become so large that they do not fit into GPU memory for available hardware. 

Table~\ref{tbl:newgraph} shows that with the new graph representation, \cop{} operates almost as quickly as the original MiniSAT while reducing the number of iterations and much faster than \emph{Graph-Q-SAT}; note that the difference is significant even though there were only at most $3$ steps made with the model.

\section{Conclusion}\label{sec:conclusion}
In this work, we have proposed an approach that limits heavy RL-based heuristics and finds the best tradeoffs between the number of iterations and their runtimes for SAT solvers. We have also introduced a novel graph representation for OSSP-based problems that greatly reduces model size and speeds up computations. In further work, we plan to extend this exciting direction, developing similar modifications for other optimization problems that can be reduced to Boolean satisfiability.

\subsection*{Acknowledgements}

This research has been supported by the Huawei project TC20211214628.

\bibliographystyle{abbrv}
\bibliography{references}
\end{document}